# A Human-Centered Workflow for Using Large Language Models in Content Analysis

Ivan Zupic
School of Creative Management
Goldsmiths, University of London
Email: i.zupic@gold.ac.uk

**Abstract**

While many researchers use Large Language Models (LLMs) through chat-based access, their real potential lies in leveraging LLMs via application programming interfaces (APIs). This paper conceptualizes LLMs as universal text processing machines and presents a comprehensive workflow for employing LLMs in three qualitative and quantitative content analysis tasks: (1) annotation (an umbrella term for qualitative coding, labeling and text classification), (2) summarization, and (3) information extraction. The workflow is explicitly human-centered. Researchers design, supervise, and validate each stage of the LLM process to ensure rigor and transparency. Our approach synthesizes insights from extensive methodological literature across multiple disciplines: political science, sociology, computer science, psychology, and management. We outline validation procedures and best practices to address key limitations of LLMs, such as their black-box nature, prompt sensitivity, and tendency to hallucinate. To facilitate practical implementation, we provide supplementary materials, including a prompt library and Python code in Jupyter Notebook format, accompanied by detailed usage instructions.

**Keywords:** Large Language Models, LLM, Content Analysis, Qualitative research, Quantitative research, Qualitative coding, Annotation, Summarization, Information extraction, Validation

**This version:** 27 February 2026.

Click here for the latest version and supplementary materials (Python code and prompt library).

# Table of Contents

# Introduction

Content analysis, defined as a research technique for making replicable and valid inferences from texts to the contexts of their use (Krippendorff, 2012), has long been central to qualitative and quantitative research in management and organization (Duriau et al., 2007). However, traditional content analysis methods face significant scalability challenges. While analyzing a handful of documents is feasible, processing hundreds of interviews or millions of social media posts exceeds human capacity (Barros et al., 2025), potentially causing important insights to be overlooked in massive datasets. These limitations have prompted growing interest in computational tools that can support content analysis without sacrificing validity (McKenny et al., 2018).

Traditional computational text analysis tools (e.g., topic modeling, sentiment analysis) have made significant progress towards enabling stronger insights from massive text data (Alaei et al., 2019; Hannigan et al., 2019; Schmiedel et al., 2018; Wetzels et al., 2025). However, Large Language Models (LLMs) offer even a more compelling solution to these scalability challenges. Unlike specialized machine learning models that have narrow capabilities and typically require extensive training for specific tasks, LLMs demonstrate remarkable versatility through their emergent abilities: capabilities that appear when models are scaled up in size and training data (Wei et al., 2022). These models can handle diverse tasks simply by understanding natural language instructions from users, making them accessible to researchers without extensive technical expertise in natural language processing or machine learning. This accessibility represents a fundamental paradigm shift: researchers can program these models for specific tasks through prompting rather than requiring bespoke algorithms for each application.

Because LLMs can perform content analysis at scale with measurable accuracy, they can significantly expand the empirical scope of content analysis research. At the same time, their use is not without challenges. LLMs are sensitive to how instructions are phrased, prone to generating plausible but inaccurate outputs, and their internal workings remain largely opaque (Chae & Davidson, 2025; Törnberg, 2024a). Adoption without appropriate methodological care risks introducing systematic errors that go undetected, with potentially serious consequences for research validity (Ashwin et al., 2025; Lin & Zhang, 2025).

The main aim of this paper is to legitimize and standardize the use of LLMs for content analysis. As such we draw on diverse literature in political science, sociology, natural language processing, information systems, psychology, management, and our own experience to develop a human-centered workflow for using LLMs for content analysis. We build on and extend previous contributions on LLM research workflows (Carlson & Burbano, 2025; Than et al., 2025; Törnberg, 2024a). Our comprehensive treatment of this process and practical resources (Python code and prompt library) offer researchers a one-stop shop for the process of using pre-trained LLMs through prompting for content analysis. While our aims are ambitious, there are also clear boundaries to what we want to do. We do not empirically compare models or validation techniques. We do not cover fine-tuning, retrieval-augmented generation (RAG) models and other advanced techniques. We are focused on providing guidelines and resources for researchers that do not require significant up-front investment or exquisite technical expertise that are beyond typical management researcher's capabilities.

This paper makes three core contributions to the emerging literature on using LLMs in management research workflows. First, it conceptualizes LLMs as universal text processing machines (as opposed to conversational partners), arguing that scholarly value comes from processing text data in programmable, API-based pipelines that can be documented, audited, and reproduced. Second, it offers an explicitly human-centered, end-to-end workflow for LLM-enabled content analysis, spanning research design, data preparation, promptbook development, scalable processing, validation/reliability/robustness assessment, and interpretation. This integrates and systematizes fragmented guidance across diverse fields, such as management, politics, sociology, computational social science, and natural language processing (NLP). Third, it advances methodological standards for publishable LLM-assisted research by (a) organizing the main use cases into three task families (annotation, summarization, and information extraction), (b) operationalizing codebooks as "promptbooks" with structured outputs to improve traceability, and (c) prioritizing validity, reliability, and robustness checks (including gold-standard benchmarking where possible and stability/sensitivity analyses where not). Collectively, these contributions bridge qualitative and quantitative traditions in content analysis while providing practical, replicable resources (prompt library and code templates) that lower barriers to rigorous, transparent adoption.

## Content analysis

Content analysis is broadly defined as "any methodological measurement applied to text (or other symbolic materials) for social science purposes" (Shapiro & Markoff, 1997, p.14, as cited in Duriau et al., 2007; Reger & Kincaid, 2021). More precisely, Krippendorff (2012) defines it as "a research technique for making replicable and valid inferences from texts (or other meaningful matter) to the contexts of their use." As a superordinate category, content analysis encompasses a wide variety of analytic approaches, ranging from purely qualitative methods used in grounded theorizing, literature reviews and case-based research, to highly quantitative techniques such as dictionary-based coding, topic modeling, and natural language processing (Reger & Kincaid, 2021). Central to the method is the assumption that textual analysis allows researchers to access the cognitive schemas, values, intentions, and attitudes of those who produced the text, structures that are often difficult to study through conventional quantitative means (Duriau et al., 2007; Krippendorff, 2012).

Content analysis offers several notable advantages for organizational research. Because it can be applied to documents and communications that organizational actors produce in the ordinary course of their work, such as annual reports, letters to shareholders, and meeting transcripts, it provides an unobtrusive means of studying sociocognitive processes in context, thereby avoiding some of the demand-characteristic biases associated with surveys and experiments (Duriau et al., 2007; Reger & Kincaid, 2021). The method is also analytically flexible: it can be applied at the level of manifest content, capturing surface-level text statistics, or at the level of latent content, where deeper interpretive meaning is sought (Duriau et al., 2007). It can be used to convert unstructured text into structured data for further statistical analysis (Reger & Kincaid, 2021). It works well for both deductive theory-testing approaches and inductive theory-building endeavors. Because content analysis can be applied to existing archival materials, it is well-suited to longitudinal research designs and is relatively cost-effective compared to primary data collection (Duriau et al., 2007; Insch et al., 1997).

Despite these strengths, content analysis carries significant methodological demands, particularly with respect to reliability and validity. Reliability in content analysis is typically assessed through intercoder agreement, i.e., the degree to which independent coders assign the same categories to the same units of text (Insch et al., 1997; Krippendorff, 2012). Simple percentage agreement has been widely criticized as an inadequate reliability measure, as it does not correct for chance agreement. More robust alternatives, such as Krippendorff's alpha or Cohen's kappa, are therefore recommended (A. F. Hayes & Krippendorff, 2007). Validity concerns are equally important: the coding scheme must demonstrably measure the construct it is intended to capture, and categories must be semantically coherent and consistently applied (Krippendorff, 2012; Morris, 1994). To meet these standards, researchers are advised to follow a systematic, iterative coding protocol: testing, revising, and re-testing the coding scheme before applying it to the full dataset (Insch et al., 1997; Reger & Kincaid, 2021).

## Large Language Models (LLMs)

LLMs are advanced Artificial Intelligence (AI) systems designed primarily for understanding and generating human language (A. S. Hayes, 2025). Technically, LLMs are neural networks based on transformers (Vaswani et al., 2017), a specific type of deep learning architecture (LeCun et al., 2015). They are trained on massive amounts of text data. The results of this pre-training are stored in model parameters called weights, the number of which could range from millions for smaller models to hundreds of billions for larger models. Crucially, LLMs demonstrate emergent abilities: when models are scaled up in size and training data, they often exhibit new capabilities (such as zero-shot reasoning or commonsense inference) that smaller models do not have (Wei et al., 2022). This means an LLM can often handle tasks it was never explicitly trained for, simply by "understanding" an instruction from user. This property is particularly useful for content analysis research tasks, as we can prompt the model to perform tasks like annotation, summarization, or information extraction without needing a bespoke algorithm for each. This property of LLMs makes them highly useful and easy to use compared to applying traditional machine learning algorithms, which needed special training for each specific task, requiring higher input levels of technical expertise, budget and human work.

LLMs are essentially word predictors. They take text (called prompt) as an input and based on the probability distribution of words learned during training they proceed – word by word – to produce seemingly coherent and contextually relevant text (Nguyen & Welch, 2025; Ornstein et al., 2025). Yet, models have no inherent understanding of text, they are basically matching patterns learned during training.

It is important to be aware of the basic functioning of these models. Understanding these basics has practical implications. Because LLMs predict based on patterns, they perform best on tasks similar to those represented in training data (Chollet & Watson, 2025; Than et al., 2025). They excel at identifying explicit textual features but sometimes struggle with deeply contextual interpretation requiring domain expertise or cultural knowledge absent from training (Schroeder et al., 2025). The token-based processing means LLMs can miss relationships spanning long documents unless specifically prompted to consider broader context. Researchers should view LLMs as sophisticated pattern-matchers rather than autonomous interpreters. They are powerful tools that require human guidance and validation. This knowledge can make researchers less susceptible to hype and more sensitive to models' limitations while still being able to exploit models' capabilities.

We summarized essential vocabulary for LLM technology in Table 1.

| Term | Explanation |
|---|---|
| Token | A token is the basic unit of text that an LLM processes. Before text can be analyzed by an LLM, it needs to be split into tokens by an algorithm called tokenizer. In English, 1 token typically equates to 0.75 word on average. Token limits determine how much text one can input/output. API costs are often calculated per token. Context windows are measured in tokens. |
| Context window | A context window is the maximum amount of text an LLM can "see" and work with at one time. It's one of the key limitations of LLMs as they can't remember anything outside their context window. The length of documents we want to process in one API call need to be significantly smaller than the context window of the LLM used for processing. The size of a context window is measured in tokens. |
| Parameter/Weight | Parameter is the umbrella term for all LLM values learned during training. Weight is a specific type of parameter that determines how much influence one neuron in neural network has on another. Because weights form the vast majority of parameters of an LLM, the terms are sometimes used interchangeably, although technically they are not the same. |
| Inference | Inference is the process of an LLM generating a response or making a prediction based on the input provided (prompt). The process uses the trained neural network to predict what text should come next, token by token. |
| Encoder-decoder | A neural network architecture used in many transformer models. The 'encoder' processes the input text into a numerical representation (vectors), and the 'decoder' generates the output |

| | text from that representation. While original transformers used both, many modern LLMs (like GPT) are 'decoder-only' architectures optimized for text generation. Encoder–decoder models remain common in translation/summarization research. |
|---|---|
| JSON (JavaScript Object Notation) | A standard text-based format for representing structured data based on JavaScript object syntax. It is increasingly the standard output format for LLM research tasks because it forces the model to organize free-text response into a machine-readable structure, facilitating automated parsing into databases or spreadsheets. Schema-constrained outputs help reduce invalid responses. |
| API (Application Programming Interface) | A standardized way for software to send requests to an LLM and receive outputs programmatically. APIs enable batch processing, logging of inputs/outputs, settings control, and reproducible pipelines. |
| Temperature | A key configuration setting that governs the degree of randomness in how a large language model (LLM) selects its next output token. Lower values (often ~0) produce more consistent, predictable output and are preferred for measurement-like tasks (classification, extraction). Higher values can increase creativity/readability but reduce run-to-run stability and are preferrable for tasks like summarization. |

*Table 1. Essential vocabulary of LLM technology.*

Researchers typically access LLMs through two interfaces. Chat interfaces (like ChatGPT, Claude.ai) are user-friendly and useful for prompt development, testing, and analyzing small samples. However, they're impractical for processing hundreds or thousands of texts. API (Application Programming Interface) access allows programmatic interaction (Churchill et al., 2025). Researchers write code that sends texts to the LLM and receive structured responses, which are then stored in a spreadsheet or database. This enables batch processing and reproducible workflows. APIs typically charge per token processed, making cost estimation important for large projects. Processing hundreds of documents might be within reach of most researchers (especially as some models have generous free-tier allocations), while processing thousands and millions of documents requires significant budgets and careful planning (Ashwin et al., 2025; Chae & Davidson, 2025). Open-source models eliminate usage costs but require technical infrastructure and expertise for deployment (Schroeder et al., 2025).

APIs allow researchers to (1) automate processing across large corpora; (2) standardize inputs, parameters, and outputs; (3) enforce structured formats (e.g., JSON); (4) log prompts, model versions, settings, and timestamps; and (5) integrate validation and robustness checks directly into the pipeline (Carlson & Burbano, 2025; Pangakis et al., 2023; Schroeder et al., 2025). In other words, APIs turn an LLM from a conversational partner into a measurable instrument embedded in a documented research procedure. The API-based workflow resembles a classic content-analysis pipeline: researchers define units of analysis and a codebook, translate that codebook into a "promptbook," run batch inference, and then validate the resulting measures against human judgments or other evidence (Stuhler et al., 2025).

A useful conceptual frame is thinking of LLMs as capable but literal research assistants who excel at following detailed instructions but lack independent judgment. Like a well-trained RA, LLMs can consistently apply coding schemes across thousands of texts, identify relevant passages, and organize information, but they require explicit guidance, quality control, and human oversight. They won't catch conceptual inconsistencies in researcher's coding scheme or question whether the research design makes sense. This metaphor clarifies appropriate delegation: use LLMs for systematic, rule-based processing while retaining human responsibility for theoretical interpretation, methodological decisions, and validity assessment. With this metaphor we do not wish to anthropomorphize, but to highlight an epistemic stance: LLM outputs are candidates for evidence that must be verified, not findings that speak for themselves.

**LLM Models**

In applied research, "the model" is a specific released system (e.g., a particular GPT or Gemini variant: "gpt-5.2-2025-12-11", "gemini-2.5-flash") with a defined context window, model size (i.e., the number of parameters), pricing, latency, and safety/guardrail behavior. These properties matter because they shape what texts researchers can feed in (context window), model capabilities (size), how expensive it is to process the corpus (pricing), how quickly one can run robustness checks (latency/rate limits), and whether the model will refuse or rewrite some content (safety policies).

The choice of model dictates the 'reasoning' capability available for qualitative tasks. For instance, smaller models might suffice for simple binary classification (e.g., positive/negative sentiment), whereas complex deductive coding requiring nuance and cultural context often necessitates larger, state-of-the-art models to minimize reasoning errors.

LLM outputs are stochastic by default. Running the same prompt on the same model two times might yield different results. However, API access allows researchers to set certain parameters that influence the randomness of outputs. As LLMs generate text token by token, parameters such as temperature (randomness of token sampling) and top-p/top-k (how broadly the model samples from likely next tokens) influence both accuracy and reliability. For measurement-like tasks (classification, extraction), researchers should generally prefer low-variance settings (often temperature ≈ 0) and then explicitly test stability by re-running subsets and checking agreement. For generative tasks (summarization), slightly higher randomness can sometimes improve readability, but it increases run-to-run variation and should be paired with stability checks and/or deterministic post-processing. Model choice and parameter settings are part of operationalization and must be reported like any other measurement decision (Foisy et al., 2025; Watkins, 2024).

When selecting a model for content analysis, researchers must navigate tradeoffs between model size, reasoning capabilities, and cost as well as between proprietary (closed) and open-source models. Same providers typically often multiple models of varying capabilities. Large proprietary models offer superior zero-shot reasoning but introduce data privacy risks and per-token costs. Smaller open-weights models (e.g., Llama) allow for local, private processing and zero marginal cost but may require computing infrastructure, significant technical expertise, additional fine-tuning or few-shot prompting to match the accuracy of proprietary counterparts on complex annotation tasks (Yang et al., 2025).

Model selection involves multiple tradeoffs:

- Capability vs. cost/speed. Flagship models (e.g., GPT-5.2, Gemini 3) perform best but are expensive and often slower. Smaller models sometimes suffice for straightforward tasks at fraction of the cost (Alizadeh et al., 2024). While costs are decreasing, analyzing massive datasets (e.g., millions of documents) with flagship models can still be prohibitively expensive (Ashwin et al., 2025).
- Speed vs. accuracy. Larger models produce higher-quality outputs but process texts more slowly. This is important when coding thousands/millions of documents.

- Knowledge cutoff dates vary by model, affecting analysis of recent events or emerging concepts.
- Context window size determines whether one can process entire documents or must segment them and process in chunks. Researchers should pilot multiple models on representative samples before committing to large-scale processing. Performance differences can be substantial for complex tasks but minimal for simple classification.

**Practical aspects of using LLMs**

The beauty of using LLMs is that we can use them for processing different research tasks just by changing the prompt. While LLMs are often not the best (aka the fastest, most accurate) tool for many text processing tasks (Brandt et al., 2025), they are very simple and versatile to use and will thus appeal to many researchers without exquisite technical knowledge and budget to support more sophisticated NLP workflows.

Advantages and challenges of using LLMs for content analysis are summarized in Table 3.

| Advantages | Challenges |
|---|---|
| (1) Scalability. Process thousands of texts in hours vs. months of human coding | (1) Validation burden. Each task requires systematic accuracy assessment. |
| (2) Consistency. Apply identical criteria across entire dataset without coder fatigue. | (2) Cost uncertainty. API expenses can escalate with large datasets. |
| (3) Transparency. Prompt documentation provides explicit audit trail of decision rules. | (3) Dependency on external systems. Proprietary models may change or be discontinued. |
| (4) Flexibility. Rapidly test alternative coding schemes or theoretical frameworks, | (4) Limited contextual understanding. LLMs may miss cultural nuances or organizational specifics humans would recognize, especially for topics underrepresented in training data. |
| (5) Augmentation. Free researchers to focus on theoretical interpretation rather than mechanical coding | (5) Replication barriers. Exact reproduction is difficult due to model updates and nondeterminism. |

*Table 2. Advantages and challenges of using LLMs for research.*

Researchers accessing LLMs via API will encounter rate limits (requests per minute/RPM and tokens per minute/TPM). Exceeding these results in errors, requiring code that implements waiting times between API calls. Pricing is typically charged per 1M input/output tokens. Flagship models typically cost more per token than older or smaller models. Costs are generally decreasing over time. However, analyzing millions of documents with flagship models remains a significant line item that must be budgeted for in grant applications. Rate limits, batching, and cost monitoring also shape feasible validation and robustness. Rate limits constrain how often researchers can re-run prompts for intra-prompt stability, inter-prompt stability, or inter-model agreement. Pricing determines whether researchers can afford best practice (e.g., multiple models, multiple prompt variants, repeated runs).

Because LLMs are trained on internet data, they may have already "seen" the text researchers are asking them to annotate (e.g., academic articles, social media posts, annual reports). This "data contamination" can artificially inflate performance metrics, making the model appear more capable than it is when applied to truly novel data (Pangakis et al., 2023).

**Ethical and legal considerations**

The use of LLMs in content analysis raises several ethical and legal issues that researchers must address.

Proprietary LLM APIs require transmitting data to external servers, which raises important privacy concerns. Data shared through web browser or API access points is shared with the provider company, and researchers working with sensitive participant data must carefully weigh this risk (Schroeder et al., 2025). Researchers handling confidential organizational data, personally identifiable information (PII), or data governed by regulations such as GDPR or HIPAA face particular challenges, as data submitted to platformed LLMs such as ChatGPT is likely to be used for model training (Törnberg, 2024a). GDPR compliance may, for instance, require signing a formal Data Processing Agreement with the service provider (Törnberg, 2024a). For sensitive data, researchers should consider several protective strategies: (1) using open-source models deployed locally to maintain full data control, since these models are designed so that data need not be shared with any third-party entities (Alizadeh et al., 2024; Schroeder et al., 2025); (2) anonymizing or de-identifying data before API transmission, though this may reduce analytical value (Törnberg, 2024a); (3) obtaining explicit organizational approval for third-party data processing (Lin & Zhang, 2025); or (4) using API agreements or model configurations that prohibit training on customer data, which is a common opt-out practice when using proprietary providers (Carlson & Burbano, 2025; Lin & Zhang, 2025).

Attribution and intellectual property questions arise when LLMs generate substantial portions of coding schemes or analytical outputs. Researchers have noted significant uncertainty about whether AI-assisted creative outputs carry intellectual property implications, and whether relying on LLM-generated analysis constitutes a form of accidental academic plagiarism (Schroeder et al., 2025). While the broader research community has yet to reach consensus on these questions, transparency is widely recommended: disclosures should ideally detail which models were used, any prompt engineering performed, and how LLM involvement shaped the analysis (Chew et al., 2023; Schroeder et al., 2025).

Bias and fairness considerations extend beyond technical validation to encompass representational harms. LLMs may systematically misinterpret or misrepresent texts from underrepresented demographic groups, as their training on large-scale scraped text data tends to reflect majority, often Western, perspectives (Abdurahman et al., 2025; Churchill et al., 2025). LLMs tend to misportray marginalized groups and flatten within-group heterogeneity (Lin & Zhang, 2025). Ashwin et al. (2025) demonstrate empirically that LLM annotations can introduce systematic bias when applied to qualitative data from contexts not well represented in standard training corpora. Researchers should therefore assess whether their constructs and coding schemes embed culturally specific assumptions that may not transfer across contexts, particularly when working with non-Western or otherwise underrepresented organizational settings (Abdurahman et al., 2025; Ashwin et al., 2025).

## LLMs as Universal Text Processing Machines

For this paper we conceptualize Large Language Model (LLM) as a **universal processing machine for text data, programmable by prompt (**Figure 1**)**.

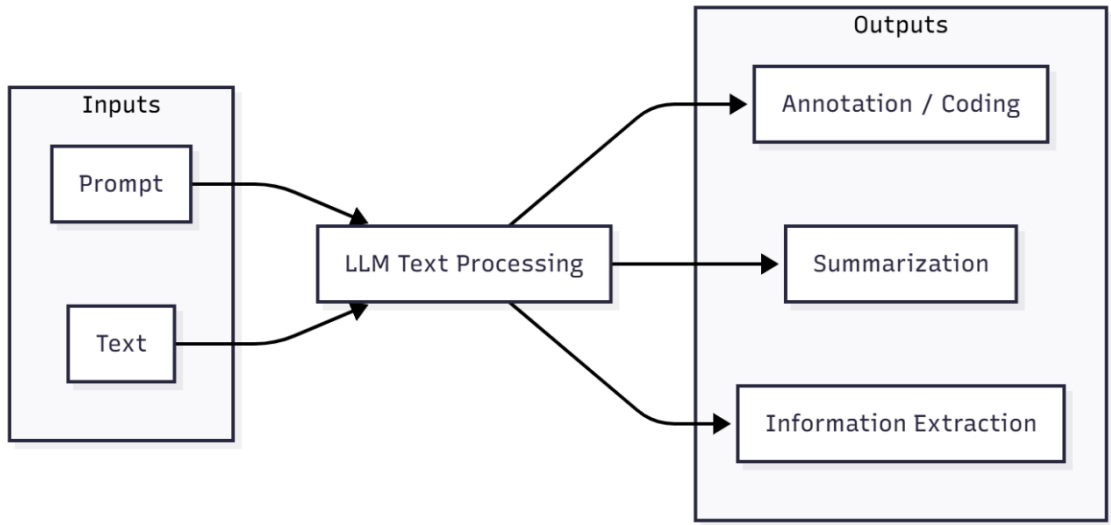

*Figure 1. LLMs conceptualized as universal text processing machines.*

LLM is a flexible analytical instrument that takes two inputs (text data and a researcher-defined prompt) and transforms them into structured analytical outputs. What makes this framing powerful is the notion of universality: by simply changing the prompt, the same underlying model can be directed to perform fundamentally different content analysis tasks. These tasks are organized into three categories that span the main demands of content analysis research: **annotation** (classifying or coding text into theoretically meaningful categories), **summarization** (condensing text into concise, targeted representations), and **information extraction** (identifying and pulling out specific, granular data points embedded in the text). The machine metaphor positions the LLM as a controlled measurement instrument, analogous to a human coder or a dictionary-based classifier, whose outputs are systematically reproducible, auditable, and subject to the same standards of validity and reliability that rigorous content analysis has traditionally demanded. The focus is on prompt design, positioning the promptbook as the primary instrument through which researchers operationalize their constructs and control the analytical process. This framing anchors the entire workflow: the LLM is a machine and the researcher's job is to engineer the instructions that program it, validate its outputs, and retain interpretive authority throughout.

**Annotation (Qualitative coding/text classification)**

We use text annotation as an umbrella term for the task of assigning text to categories (codes, labels, annotations). The process can be called different names in different literatures, epistemological approaches and disciplines: annotation, labeling, qualitative coding or text classification. While there is large overlap, there are also subtle differences and disciplinary conventions.

Annotation, is the creation of structured scheme of variables from unstructured text (Carlson & Burbano, 2025). The goal of **qualitative coding** is to find themes in text content, understand meaning and build new theory. LLMs can assist with qualitative coding in both inductive and deductive approaches. LLMs can facilitate deductive coding where labels are generated a priori based on existing hypotheses or previous research. They can apply existing codes to large datasets rapidly (Scheuerman & Acklin, 2024). On the other hand, LLMs can also help in inductive, codebook emergent approaches. **Text classification** is a canonical task in Natural Language Processing (NLP). This term is often used by computer scientists. Its purpose is to automatically sort data into predefined or emergent categories, which are typically mutually exclusive and systematic.

Use cases of LLM annotation applications in management research include:

- Literature review screening. Classifying thousands of abstracts by relevance to inclusion criteria, reducing manual screening burden.
- Coding qualitative data (e.g., interviews, annual reports, social media posts, newspaper articles, …) according to a predefined schema.
- Sentiment analysis. Coding stakeholder communications, online reviews, or social media for sentiment (positive/negative/neutral) toward organizations or products.
- Theoretical construct identification. Detecting presence/absence of constructs like organizational learning, entrepreneurial orientation, or leadership styles in documents.
- Regulatory compliance. Categorizing corporate disclosures by regulatory requirement or identifying risk factor themes in 10-K filings.

**Summarization**

The task of summarization involves condensing long-form textual information into a short, concise, and coherent summary while retaining the essential meaning or capturing specific aspects of the original text (Arora et al., 2025). This capability is a core function of Large Language Models (LLMs) and is highly valuable for various research applications.

**Extractive summarization** selects and combines existing sentences or phrases directly from the source text, essentially highlighting the most important passages. **Abstractive summarization** generates new text that conveys the source's meaning using different words and sentence structures. This section primarily covers abstractive summarization, while extractive summarization will be the topic of information extraction section.

Researchers are not limited to summarizing whole documents but can aim to summarize specific aspects of text. While **generic summarization** aims to capture the overall content without a specific angle, **query-focused or targeted summarization** emphasizes information relevant to a particular question or topic. Targeted summarization significantly expands the use cases for the use of summarization in research workflows.

Summarization enables several research workflows, for example:

- Interview analysis. Condensing lengthy interview transcripts into structured summaries organized by themes.
- Meeting minutes. Extracting decisions, action items, and key discussion points from recorded meetings or ethnographic field notes.
- Literature synthesis. Generating structured summaries of papers organized by theory, methods, findings, and contribution, which is useful for systematic literature reviews.
- Temporal tracking. Creating consistent summaries of organizational communications (earnings calls, press releases) across time to track strategic narrative evolution.
- Labeling results of topic modeling.

**Information extraction**

Information extraction (IE) is the task of automatically identifying and extracting specific, granular pieces of information embedded within texts (Stuhler et al., 2025). Unlike summarization, which condenses content while preserving narrative flow, information extraction targets specific data points, entities, relationships, or facts and organizes them into a structured format (e.g. a database or JSON object).

Different types of information can be extracted with IE (Xu et al., 2024). **Named Entity Recognition (NER)** focuses on extracting substrings within a text that name real-world entities (e.g. companies, persons) (Keraghel et al., 2024). **Relation Extraction (RE)** tries to extract relations between entities (e.g., [person] is a CEO of [company]; causal relationships between constructs). The extracted relations can populate knowledge graphs or relational databases. Researchers can also use LLMs to identify and record specific events and temporal metadata (e.g., acquisition by [buyer] of [target] at [date]). Information on entity attributes (e.g. information about product features, materials or price) or specific text snippets focusing on the topic of interest can be extracted. Researchers can constrain a LLM to output specific phrases, entities, or sentences exactly as they appear in the original source text, without paraphrasing, normalization, or synthesizing new content.

LLMs have made prompt-based IE highly accessible and flexible, allowing researchers with basic coding skills to define their own tasks and extract relevant information without requiring specialized NLP models or extensive training data (Stuhler et al., 2025).

Management researchers can leverage IE for:

- Corporate governance analysis. Extracting board composition, executive backgrounds, and committee structures from proxy statements across thousands of firms
- Event studies. Identifying specific events (mergers, acquisitions, leadership changes, partnerships) with precise dates and entities from news archives.
- Survey analysis. Extracting structured themes from open-ended survey responses while preserving respondent language.
- Meta-analysis. Systematically extracting relationships between constructs, effect sizes, sample characteristics, and methodological details from published papers.

## A Human-Centered Workflow for Using LLMs

This workflow is based on extensive methodological literature from social science disciplines, notably Political science and Sociology. The process can be applied to different kinds of text data, from academic articles and annual reports to social media data (e.g., Reddit, X). The process effectively converts a corpus of unstructured text into a structured table where each column is the result of specific application of one of tasks: annotation, summarization or information extraction. The input to this process is text data stored either as file collection or in tabular format (e.g. Excel). The output is a structured table with columns representing annotations, summaries and extracted information. This newly created dataset can then be used in further quantitative or qualitative analysis.

There are six stages in this workflow: (1) Research design, (2) Data preparation, (3) Promptbook development, (4) Processing, (5) Assessing validity, reliability and robustness and (6) Interpretation and further analysis. They are summarized in Figure 2.

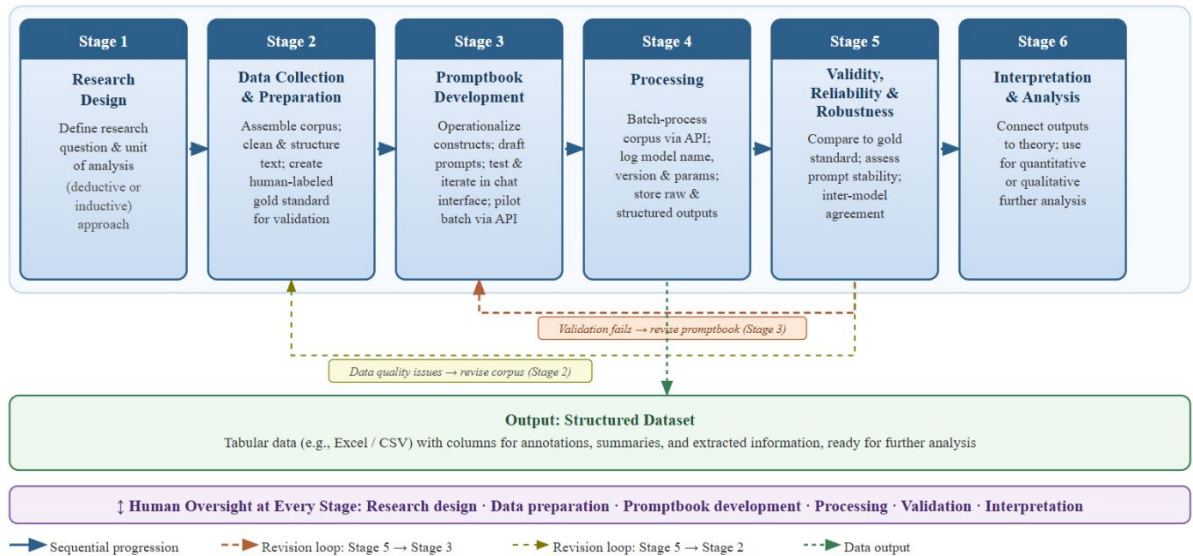

*Figure 2. A Human-Centered Workflow for Using LLMs in Content Analysis.*

## 1. Research design

Appropriate research questions for LLM-augmented content analysis generally fall into categories where the task involves transforming unstructured text into structured data, classifying content based on specific definitions, or extracting explicit information. The literature suggests that LLMs are most effective when the task relies on explicit textual evidence and clear definitions, while tasks requiring deep cultural inference or interpretation of highly subjective concepts may yield less reliable results (Stuhler et al., 2025; Yang et al., 2025). LLMs are most appropriate when research requires: (1) systematic categorization or measurement of textual features, (2) processing volumes exceeding practical human capacity, (3) consistent application of explicit coding criteria, and (ideally) (4) tasks where validation against ground truth is feasible.

However, some research questions are less appropriate for LLM use. Some specific conditions where LLM analysis is risky or prone to failure include research requiring deep cultural nuance, non-normative communication or identity-based inference. LLMs may struggle with "navigational capacity" or complex aspirations in non-Western contexts (e.g., Rohingya refugees), where they introduce bias based on the subject's demographics rather than the text content (Ashwin et al., 2025; Pavlovic & Poesio, 2024). LLMs have demonstrated an inability to accurately interpret the lived experiences and communication styles of people with intellectual and developmental disabilities, often failing to capture the "truth-value" of their statements (Friedman et al., 2025). Models may hallucinate or improperly infer demographic details (e.g., religious affiliation) based on names or nationalities, even when explicitly instructed not to (Stuhler et al., 2025). LLMs might also struggle with analyzing texts in languages or domains with limited training data representation.

Researchers need to translate the research question into measurement tasks. They should specify: (1) the unit of analysis (document, paragraph, sentence, turn-at-talk) (Chae & Davidson, 2025); (2) whether the approach is deductive (codebook-first) (Chew et al., 2023) or inductive (codebook-emergent) (Garcia Quevedo et al., 2025; Ziems et al., 2024); (3) the intended use of outputs (descriptive, predictive analysis, theory building) (Carlson & Burbano, 2025; Ziems et al., 2024); and (4) the validity argument for using LLM outputs as indicators of constructs (Halterman & Keith, 2025). This step also clarifies what not to ask the model to do: interpretive claims (e.g., "what the participant really meant") should remain a human analytic responsibility, even if LLMs assist with organizing evidence.

## 2. Data collection and preparation

The next step is assembling a corpus that is both representative and suitable for automated processing. Researchers typically draw their data from sources such as media articles, social media posts, annual reports, academic papers, digitized archival documents, open-ended survey responses, or administrative records, all of which exist as unstructured text rather than tidy numeric datasets (Schwitter, 2025). Because the quality of downstream results depends heavily on the quality of the input, careful thought must be given to data coverage and sampling strategy from the outset. For larger datasets, stratified random sampling (using available covariates to ensure representativeness across relevant subgroups) is preferable to simple random sampling (Churchill et al., 2025).

Once a corpus has been assembled, the data must be cleaned and structured before it can be reliably processed. Text data collected from real-world sources is frequently messy: social media posts contain misspellings and non-standard language, while digitized historical documents often suffer from optical character recognition (OCR) errors that introduce artifacts into the text (Schwitter, 2025). LLMs have shown a notable ability to handle such irregularities, correcting OCR errors and extracting key information even from poorly formatted source material, though the principle of "garbage in, garbage out" still applies: particularly difficult or corrupt inputs will produce less reliable outputs (Schwitter, 2025). For fine-tuning or few-shot workflows, text must additionally be formatted according to the requirements of the chosen model architecture, with inputs and expected outputs structured consistently across all training examples (Alizadeh et al., 2024).

If feasible, researchers should also create a human-labeled development and evaluation sets, which serve as the ground truth against which model performance is evaluated, and prompt strategies are refined. Researchers can use this approach both to test different prompting approaches and to measure final model accuracy (Alizadeh et al., 2024; Lin & Zhang, 2025). The development set and a separate evaluation set should be sampled independently and randomly from the broader corpus to avoid overlap, since using the same cases for both prompt optimization and performance reporting risks inflating reported accuracy (Lin & Zhang, 2025). Crucially, the human content analysis process itself must be grounded in a clear codebook that defines each category, provides examples, and specifies how to handle ambiguous cases. Without this, there is a conceptual gap between what human coders and the LLM are actually measuring, which threatens the validity of any comparison between them (Pangakis & Wolken, 2024; Törnberg, 2024a). It is worth acknowledging, however, that human annotations are not infallible ground truth: inter-coder agreement on complex tasks is often lower than expected, and in some cases LLMs can equal or surpass the consistency of human expert coders (Bermejo et al., 2025; Bisbee & Spirling, 2025; Törnberg, 2024b).

## 3. Promptbook development

A prompt is a text-based input that is fed to a LLM to guide its output, with the primary purpose of providing the language model with instructions and context for achieving a desired task (Marvin et al., 2024). It is typically a mix of instructions, questions, examples ("demonstrations"), formatting requirements, or other constraints that guide the model's output (Chang et al., 2024). Researchers are advised to operationalize concepts, i.e. convert abstract research questions into a detailed codebook or "promptbook" with definitions, rules, and examples (Stuhler et al., 2025).

The term "promptbook" draws a deliberate historical parallel to theatrical production. In theater, the promptbook is the master script containing not only the dialogue but also the technical cues, lighting instructions, and stage directions that ensure a consistent performance across different nights and casts (Stuhler et al., 2025). Similarly, in LLM-assisted research, the promptbook serves as the master protocol. It contains specific task instructions, contextual definitions, exclusion criteria, and formatting constraints that ensure the 'performance' (of annotation, summarization or extraction) remains consistent across thousands of documents. Just as a stage manager relies on the promptbook to maintain fidelity to the director's vision, the researcher relies on the promptbook to maintain fidelity to the research design.

Developing an effective promptbook is an iterative process that requires repeated testing and refinement. When prompts lack sufficient clarity or detail, the model is likely to produce inaccurate or superficial responses (Boonstra, 2025). Promptbook development itself is best understood as a cycle of drafting, evaluating, and revising until performance reaches an acceptable standard (Schulhoff et al., 2025). We recommend the following workflow.

**Step 1: Draft the output structure.** Begin by defining the concepts you want to capture and designing a draft output table based on the study's research aims and questions. Each column should correspond to a distinct analytical dimension that the model will be asked to address.

**Step 2: Develop the promptbook.** Write a draft promptbook in which each column has a clearly specified task instruction and an associated variable name. Prompts should be unambiguous, precise, and constructive, guiding the model toward desired behaviors rather than specifying what to avoid (Khalid & Witmer, 2025).

**Step 3: Iterative testing in a chat interface.** Test different versions of the promptbook in a chat interface across a small number of documents that ideally represent the diversity of the full corpus. Where possible, testing across multiple models (both within and across providers) is advisable, as performance can vary considerably (Boonstra, 2025). An important practical consideration is context window management: when applying the prompt to a new document, the context window must be reset, since prior text may otherwise contaminate the model's response. The simplest way to do this is to begin a new chat session for each document. LLMs themselves can help with promptbook development. It is often useful to ask LLM to improve the prompt and/or suggest additional variables to annotate/summarize/extract. If output quality is unsatisfactory, return to Steps 1 and 2 and revise accordingly. This cycle should be repeated until the model's responses are consistently adequate.

**Step 4: Pilot API batch processing.** Once the prompts perform well in the chat interface, proceed to test batch processing via the API on a small random sample of documents (e.g., 10–20). Formally validate the batch output using appropriate reliability and validity metrics (see the section on assessing validity, reliability, and robustness). If performance falls short, return to Steps 1–3 with particular attention to the tasks and document types that failed validation. Repeat until validation metrics reach satisfactory levels.

Close human oversight at each step is essential for ensuring that automated outputs align with the researcher's conceptual intentions (Dunivin, 2025; Pangakis et al., 2023). The result of this process is a promptbook that can be used for processing the whole corpus.

**Designing a well-structured prompt**

Several broad approaches to prompting have emerged, each with distinct strengths and tradeoffs. The most fundamental distinction is between zero-shot and few-shot prompting. In zero-shot prompting, the model is given only a task instruction with no examples, relying entirely on its pre-trained knowledge to generate a response (Boonstra, 2025; Schulhoff et al., 2025). Few-shot prompting, by contrast, includes a small number of examples in the prompt, which can meaningfully improve accuracy, though the selection of examples matters considerably, as non-representative examples risk skewing outputs (Carlson & Burbano, 2025). Few-shot prompting is also impractical to implement in long prompts that simultaneously perform many different annotation/summarization/extraction tasks. A related approach is role-based prompting, in which the model is assigned a specific identity or expertise, such as that of a domain expert, to focus its responses on the specified domain and improve relevance (Lin & Zhang, 2025; Schulhoff et al., 2025).

Another widely used technique is chain-of-thought (CoT) prompting, which instructs the model to articulate intermediate reasoning steps before arriving at a final answer. This has been sometimes shown to improve performance on complex tasks and provides a degree of interpretability by making the model's reasoning visible (Boonstra, 2025; Polat et al., 2025; Törnberg, 2024a). However, other research points towards CoT prompting not offering advantage over other prompting strategies (Tseng et al., 2025). Many modern reasoning-tuned models already do internal multi-step reasoning by default, making basic CoT less necessary and sometimes counterproductive.

More advanced approaches include self-consistency, which aggregates outputs across multiple independent runs to improve reliability, and tree-of-thought prompting, which explores branching reasoning paths before committing to an answer, though both come at higher computational cost (Carlson & Burbano, 2025; Lin & Zhang, 2025). For most content analysis tasks in management research, zero-shot or few-shot prompting combined with clear task instructions and role-based prompting represents a practical and well-supported starting point (Dunivin, 2025; Törnberg, 2024a). Simple parsimonious techniques are preferred over complex more verbose ones provided performance is comparable. Researchers can consult prompt engineering guidelines and research surveys to get deeper insight (Boonstra, 2025; Chang et al., 2024; Marvin et al., 2024; Polat et al., 2025; Sahoo et al., 2025; Schulhoff et al., 2025).

The essential components of a well-structured prompt include:

- **Role Specification.** Instructing the model to adopt a specific persona (e.g., "You are a management academic" or "You are an expert in Leadership") can prime it for the task (Pavlovic & Poesio, 2024; Than et al., 2025).
- **Task Directive.** A clear, unambiguous command that outlines the core objective of the task.
- **Context and Definitions.** This section operationalizes the concepts of interest by providing detailed definitions, inclusion/exclusion criteria, and examples for handling ambiguous cases, drawn directly from the research codebook (Dunivin, 2025; Halterman & Keith, 2025).
- **Response Formatting.** Specifying the desired output structure (e.g., a single label, a number, table row, or a JSON object) is crucial for ensuring consistent results that can be parsed and stored in a database, typically in a tabular form (e.g., Excel sheet) (Carlson & Burbano, 2025; Stuhler et al., 2025).
- **Built-in triangulation.** Good prompt design involves building the prompt in a way that enables initial triangulation within the results before checking the original documents. LLMs can be asked to explain their decisions and/or copy verbatim text from documents relevant for specific annotation/summary/information (Chew et al., 2023).
- **Missing information instructions.** Explicitly instruct the model what to do if the requested information is not found in the text (e.g. "Return N/A if none.").

A sample prompt that demonstrates these principles is provided in Appendix A. The prompt library linked in the Appendix showcases other prompts we have used for LLM processing in research projects.

**JSON (JavaScript Object Notation)** (Bray, 2017) has become the de facto standard for structured LLM outputs. This recommendation stems from its ability to ensure consistency, machine readability, and reduced errors compared to unstructured text outputs (Boonstra, 2025). The primary technical advantage of using JSON is that it forces the LLM to generate output in a standardized, structured format, which is essential for systematic data processing (Foisy et al., 2025). To fully automate the content analysis process, the model's output must be reliably readable by a computer. JSON is machine-readable using standard parsing libraries, allowing a script to process the text into a data structure (like an Excel table) for further analysis (Dunivin, 2025). Returning data in JSON ensures that it is in the same style every time. Prompting for a JSON format forces the model to create a structure and limit hallucinations (Boonstra, 2025). JSON can support complex information extraction tasks as it is capable of representing complex hierarchical entity relationships (Dagdelen et al., 2024).

Requiring verbatim extraction acts as a check against hallucination (the tendency of LLMs to generate or invent information not present in the input text). By requiring that all extracted entities be verifiable strings from the passage, researchers can mitigate this risk (Dagdelen et al., 2024; Farjam et al., 2025).

## 4. Processing

Processing operationalizes the promptbook at scale. API pipelines should implement: batching, retries, timeout handling (avoiding rate limits errors when short texts are processed very fast), and strict output parsing (e.g., reject non-JSON). Each call should be logged with a unique ID, prompt version, model name/version, parameter settings, timestamp, and the raw output. If feasible, researchers should store both (a) the structured result used for analysis and (b) the raw model response to enable later audits.

## 5. Assessing validity, reliability and robustness

One of the most important steps in the process is assessing the quality of LLM output: validity (LLM measures or extracts the concept it is intended to measure), reliability (LLM provides consistent results across repeated applications and comparable contexts) and robustness (the model performs well under slight variations in prompts, models, or data artifacts). LLMs cannot be assumed to be valid "out of the box" and ideally require validation against high-quality ground truth (Abdurahman et al., 2025; Grimmer et al., 2022; Pangakis et al., 2023). However, for some tasks (e.g., summarization, inductive qualitative coding) there is no objective ground truth, so assessing quality of results requires judgement calls.

Metrics researchers can use to check validity, reliability and robustness are summarized in Table 5.

| Metric | Definition | Use Case | Validity/Reliability/Robustness |
|---|---|---|---|

| Accuracy | Proportion of correct predictions | Overall performance on balanced datasets | Validity |
|---|---|---|---|
| Precision | True positives / (True positives + False positives) | When false positives are costly, we want to maximize precision. High precision means low false positive rate. | Validity |
| Recall | True positives / (True positives + False negatives) | When false negatives are costly, we want to maximize recall. High recall means low false negatives rate. | Validity |
| F1 Score | Harmonic mean of precision and recall | Balanced measure for imbalanced classes | Validity |
| Mean Absolute Error (MAE) | Average absolute difference from true values | Continuous outcome validation | Validity |
| Cohen's Kappa | Agreement beyond chance between raters | Inter-rater or inter-model agreement | Reliability |
| Krippendorff's Alpha | Reliability measure handling multiple raters/missing data | Complex annotation schemes | Reliability |
| Intraclass Correlation (ICC) | Consistency of continuous ratings across runs | Repeated runs of the same model with the same prompt on the same data. | Reliability |
| Prompt Stability Score (PSS) | Consistency across semantically similar prompts | Assessing prompt robustness | Robustness |

*Table 3. A summary of measures for assessing validity, reliability and robustness in LLM-augmented research.*

Validation is defined as the process of confirming that the LLM's outputs accurately measure the concepts they are intended to capture, thereby confirming the overall trustworthiness of the results (Abdurahman et al., 2025). Without validation, the results produced by the LLM cannot be accorded the status of warrantable knowledge (Nguyen & Welch, 2025). Rigorous, task-by-task validation is considered the most important step in using LLMs for content analysis and is a basic requirement for publications relying on LLMs (Törnberg, 2024a). Researchers must always validate on a task-by-task basis, as performance is tied to the specific dataset, task, and prompt design (Pangakis et al., 2023).

The cornerstone of validation is the comparison of LLM outputs to a "gold standard", a subset of the data that has been carefully annotated by human experts (Pangakis, 2023). This comparison allows researchers to quantify the model's performance using standard metrics such as accuracy, precision, recall, and F1-score. Researchers use these metrics (Bermejo et al., 2025; Stromer-Galley et al., 2025) to provide a quantitative basis for assessing whether the LLM's annotations are sufficiently aligned with human understanding for the specific research task. Researchers should not rely solely on crowd-sourced workers (e.g., MTurk) for creating the ground truth dataset, as LLMs often outperform them but may still fail to match expert quality (Bermejo et al., 2025). Using highly skilled, expert annotators (e.g., graduate students or domain experts) is preferable to create a "gold" standard dataset for validation.

Researchers should employ multiple metrics to comprehensively evaluate LLM performance (Alizadeh et al., 2024; Chae & Davidson, 2025). Accuracy measures the overall proportion of correctly classified instances. However, accuracy alone can be misleading, particularly with imbalanced datasets where models may simply predict the majority class. Precision captures how accurately the model identifies a particular class by calculating true positives divided by the sum of true positives and false positives. This indicates whether the model avoids false positives. Recall measures how many relevant examples were detected by calculating true positives divided by the sum of true positives and false negatives. This shows whether the model captures all relevant instances. F1 score provides the harmonic mean of precision and recall, offering a balanced measure that is particularly useful when comparing different approaches (Pangakis et al., 2023). For very large datasets where it is not feasible to evaluate validity for the whole dataset, researchers can compare the metrics with a small human-labeled dataset and bootstrap to predict accuracy for the whole dataset (Churchill et al., 2025). This allows researchers to estimate minimum validity measures for the whole dataset.

A fundamental challenge in the assessment of Large Language Models (LLMs) is how to conduct validation, traditionally defined as comparison against human-annotated ground truth, when such ground truth is unavailable, impractical to obtain, or arguably nonexistent for the specific task (Bisbee & Spirling, 2025). When ground truth is unavailable, validation efforts pivot toward sensitivity analysis, consistency measurement (reliability proxies), expert evaluation, and internal verification mechanisms. For example, researchers can annotate texts with multiple different LLMs (or different configurations of the same LLM) to assess the level of agreement between them.

**Inter-model agreement** is defined as the degree of agreement within one prompt across different LLMs. This concept tracks the consistency of coding decisions when comparing multiple distinct LLMs (such as comparing GPT-5.2 and Gemini 3 Pro) when they are given the same set of instructions (Than et al., 2025). It assesses whether findings generalize across LLM architectures. Its primary purpose is to assess the stability and reliability of LLM-based research findings. It helps determine if the prompt and the results derived from it are robust to details of implementation by demonstrating reproducibility using several different models. When a predetermined gold standard or ground truth for a classification task is unavailable, leveraging a multi-model analysis and measuring inter-model agreement can provide a valuable proxy for reliability and validity testing (Than et al., 2025).

Measures commonly used to quantify inter-model agreement (a measure analogous to intercoder reliability) include statistical measures like Krippendorff's alpha or Cohen's kappa. Agreement levels tend to correlate with model size: larger models (those exceeding 12 billion parameters) typically show higher agreement with one another. High agreement between models also typically correlates with human judgement (Yang et al., 2025). Comparing coding decisions between multiple LLMs is considered a further step in diagnosing prompt stability, beyond relying solely on prompt stability scores (PSS) calculated within a single model (Barrie, Palaiologou, et al., 2025).

When analyzing disagreements, researchers can qualitatively analyze why the LLM disagreed with humans. Did it find a nuance the human missed? LLMs sometimes outperform careless human coders.

Because proprietary models are opaque regarding their training data, it is difficult to determine if high validity is due to reasoning or simply memorization (data contamination). If the results are used to infer the difference between LLM and human capabilities, it is necessary to ensure the validation data was not part of the LLM's pre-training corpus, which would inflate performance metrics (memorization vs. generalization) (Abdurahman et al., 2025; Pangakis & Wolken, 2024).

Reliability refers to the consistency and reproducibility of LLM outputs. Researchers should assess two distinct forms of reliability: intra-prompt and inter-prompt stability. A key challenge with LLMs is nondeterminism: a model may produce different outputs for the same input, even with identical settings (Abdurahman et al., 2025). This necessitates specific checks for reliability and robustness. Researchers can treat the LLM as an independent coder and calculate agreement metrics between the LLM and human coders, or between different LLMs and then use Krippendorff's alpha or Cohen's kappa to measure agreement beyond chance (Abdurahman et al., 2025; Törnberg, 2024a).

**Intra-Prompt Stability** involves running the exact same prompt multiple times on the same data subset with parameters like temperature fixed. Researchers could measure the consistency of the outputs using metrics like intraclass correlation (ICC) for continuous ratings or by reporting the variation in outcomes (e.g., standard deviation, majority vote distributions) for categorical ratings (Abdurahman et al., 2025; Lin & Zhang, 2025). **Inter-Prompt Stability** measures the fragility of the prompt design by testing the consistency of classification outcomes across semantically similar but slightly different prompts. A high degree of stability indicates a robust prompt design, whereas low stability suggests the results are vulnerable to small, arbitrary changes in wording (Barrie, Palmer, et al., 2025).

**Task-Specific Considerations**

Validation is mandatory for annotation on a task-by-task basis because performance varies substantially across different classification problems (Pangakis et al., 2023). Simple binary classifications generally show higher reliability than complex multi-class or multi-label tasks (Lin & Zhang, 2025). Researchers should report precision and recall separately, as LLMs often show asymmetric performance, such as higher recall than precision (Pangakis & Wolken, 2024).

Assessment is more challenging for summarization tasks due to the inherent subjectivity in what constitutes a good summary. Researchers should combine automated metrics with human evaluation of key summary qualities such as factual accuracy, coverage of main points, and absence of hallucinations (Dagdelen et al., 2024). Manual evaluation by domain experts provides crucial validation for tasks where exact boundaries are fuzzy.

For information extraction, researchers should validate both the presence of extracted entities and their correct boundaries (Dagdelen et al., 2024; Stuhler et al., 2025). Accuracy is a good measure for extracting factual information. For structured extraction tasks, ontology-based assessment can verify whether extracted relationships align with domain knowledge and semantic constraints (Polat et al., 2025).

To support replication and allow readers to assess validity, reliability and robustness, researchers should document at least the following (Abdurahman et al., 2025):

- Complete prompts used, including system messages and formatting instructions
- Model names and specific versions with access dates
- All parameter settings (temperature, top-p, etc.)
- Validation procedures, their results and sample sizes
- Robustness checks performed

In the case that the validation, reliability and robustness checks do not yield acceptable results, researchers should return to stage 2 and amend the promptbook. This may involve providing more examples, clarifying ambiguous categories, decomposing complex tasks into simpler subtasks, or reconsidering whether the construct is amenable to LLM measurement. Iteration between validation and prompt refinement is expected and should be documented (Carlson & Burbano, 2025).

## 6. Interpretation and further analysis

Interpretation reconnects model outputs to theory and to the underlying texts. For quantitative content analysis, this involves treating LLM outputs as variables and evaluating whether inferences are robust across plausible prompts/models and consistent with validation evidence. The results produced by LLM processing will likely be just an intermediate step for further statistical analysis that will use extracted structured dataset to predict theoretically relevant outcomes.

For qualitative analysis, interpretation requires returning to the text: using LLM outputs to navigate, retrieve, and compare passages, while maintaining reflexive memoing and documenting how human judgments shaped category refinement. In both cases, the standard should be traceability: readers should be able to see how constructs were operationalized, how evidence was extracted, and where human interpretation entered the chain.

# Discussion

We develop a comprehensive, human-centered workflow for applying LLMs to three core content analysis tasks: annotation, summarization, and information extraction. We situate the workflow within the established traditions of content analysis research in the social sciences (Duriau et al., 2007; Krippendorff, 2012). In doing so, we respond to a growing methodological need: while the pace of LLM adoption in management and organizational research has accelerated dramatically, published guidance has tended to address narrow aspects of the process (e.g., prompt engineering, model selection, or validity assessment) rather than providing an integrated end-to-end account (Carlson & Burbano, 2025; Than et al., 2025; Törnberg, 2024a). Our contribution is to synthesize and systematize this fragmented guidance into a single, practically grounded workflow that researchers can follow from research design through to interpretation.

We frame LLMs as universal text processing machines that are programmable by prompt. This instrument view has important methodological consequences. It positions the LLM within a long-standing tradition in content analysis where the measurement instrument (a human coder, a dictionary-based classifier, or a supervised machine learning model) must be carefully specified, validated, and documented (Insch et al., 1997; Krippendorff, 2012). From this perspective, the promptbook is the functional analogue of a codebook: it specifies the operationalization of constructs, provides inclusion and exclusion criteria, and establishes a transparent audit trail of decision rules (Stuhler et al., 2025). Treating prompts as codebooks also enables researchers to draw on the well-developed literature on codebook design, inter-rater reliability, and construct validity when evaluating LLM outputs (A. F. Hayes & Krippendorff, 2007; Pangakis et al., 2023). This framing directly responds to the concern raised by Nguyen and Welch (2025) that LLMs cannot interpret meaning: our position is that LLMs need not interpret meaning autonomously. They perform a bounded, well-specified text transformation designed, supervised, and validated by human researchers who remain responsible for construct definition, theoretical interpretation, and the ultimate warrant attached to research findings.

The six-stage workflow (research design, data preparation, promptbook development, processing, validity and reliability assessment, and interpretation) integrates this instrument view into a coherent research practice. Each stage assigns a distinct role to human judgment. Researchers specify constructs and sampling strategies; they author the promptbook and evaluate pilot outputs; they design validation procedures and assess whether reliability thresholds have been met; and they connect model outputs to theory during the final interpretive stage. This division of labor reflects a pragmatic position on the capabilities and limitations of current LLMs. Models excel at applying well-specified categorical distinctions at scale with consistency that no team of human coders could match (Bermejo et al., 2025; Gilardi et al., 2023). However, they are sensitive to prompt wording (Barrie, Palaiologou, et al., 2025), prone to producing plausible but incorrect outputs particularly in unfamiliar or low-resource domains (Ashwin et al., 2025), and their internal reasoning remains opaque (Thomas et al., 2026). The workflow mitigates each of these weaknesses: promptbook refinement addresses sensitivity; systematic validation detects inaccuracy; thorough documentation ensures transparency. Critically, the workflow's iterative structure ensures that human oversight is a genuine quality control mechanism embedded in the research process (Abdurahman et al., 2025; Pangakis & Wolken, 2024).

This paper is explicitly discussing the task-specificity of LLM performance (Thomas et al., 2026). The literature makes clear that LLM suitability varies considerably across tasks, constructs, and datasets, and that performance established in one context cannot be assumed to transfer to another (Lin & Zhang, 2025; Pangakis et al., 2023).

The practical resources accompanying the paper: the prompt library and Python code in Jupyter Notebook format, lower the barrier to entry for researchers who are not specialists in computational methods. This matters because a recurring challenge identified in the literature is that methodological guidance is often written at a level of abstraction that makes implementation difficult for researchers without extensive NLP expertise (Chew et al., 2023; Foisy et al., 2025). By providing working code and annotated prompt templates, we enable researchers to engage with the workflow concretely rather than solely at the level of principle. The supplementary materials also serve a transparency function: by making the prompts and processing pipeline available, researchers who follow this workflow have the materials necessary to satisfy emerging standards for replication in LLM-augmented research (Abdurahman et al., 2025; Barrie, Palmer, et al., 2025).

The implications for management and organizational research are significant. Text data is abundant in organizational contexts (annual reports, meeting transcripts, survey open-ends, social media, archival documents) yet the analytical bottleneck of manual coding has historically limited the scope and scale of content analysis in the field (Duriau et al., 2007). By providing a principled workflow for applying LLMs to this data, the paper opens the door to studies that would be impractical with human coding alone. These possibilities represent a genuine expansion of the empirical scope of management research, provided they are pursued with the methodological care outlined in this workflow.

**The human in human-centered workflow**

The human-centered framework advanced in this paper positions LLMs as powerful augmentation tools rather than autonomous analytical agents. This positioning reflects epistemological commitments about knowledge production in organizational research. LLMs excel at pattern matching and categorization but lack the contextual understanding, theoretical insight, and reflexive judgment essential to rigorous academic inquiry (Nguyen & Welch, 2025; Schroeder et al., 2025).

Human judgment remains essential at every workflow stage. In research design, researchers determine which questions merit investigation and how abstract constructs should be operationalized. During promptbook development, they encode theoretical knowledge into classification schemes and provide examples that reflect conceptual nuance. Validation requires human assessment of whether LLM outputs capture intended meanings or merely surface patterns. Interpretation reconnects model outputs to theory, contextualizes findings within broader literature, and generates theoretical insights that transcend pattern description.

This division of labor leverages complementary strengths. LLMs process vast textual corpora with consistency impossible for humans, but humans bring theoretical knowledge, domain expertise, and capacity for reflexive insight. The workflow enables researchers to focus cognitive effort where it matters most: conceptualization, interpretation, and theory building, while delegating repetitive coding to computational systems.

However, the human-centered approach demands new skills. Researchers must develop competence in prompt engineering, understand LLM capabilities and limitations, design appropriate validation strategies, and maintain critical distance from seemingly authoritative outputs. Graduate training programs should integrate these competencies into research methods curricula while preserving emphasis on theoretical development and interpretive sophistication.

**Future research directions**

One of the most pressing methodological challenges for future work concerns the foundations of validation itself. The workflow presented in this paper, like most current guidance in the literature, rests on the assumption that human-annotated data provides a reliable gold standard against which LLM outputs can be evaluated. This assumption is increasingly difficult to sustain. LLMs are already approaching or surpassing human expert performance for a growing range of text analysis tasks, and the trajectory of improvement is monotonically upward (Bermejo et al., 2025; Bisbee & Spirling, 2025; Törnberg, 2024b). When LLM accuracy equals or exceeds that of the human coders used to construct a validation set, the conventional validation logic (measuring the gap between machine and human judgment) becomes conceptually incoherent.

Future research must therefore develop validation frameworks that do not depend on human annotations as the unambiguous ground truth. Promising directions include sensitivity analysis approaches that assess the robustness of downstream inferences to plausible variation in coding decisions (Bisbee & Spirling, 2025), ensemble methods in which disagreement between multiple LLMs serves as a proxy for construct ambiguity rather than model error (Barrie, Palmer, et al., 2025; Than et al., 2025), and statistical bounds that characterize the range of conclusions consistent with a given level of measurement uncertainty (Carlson & Burbano, 2025). More fundamentally, the field needs to confront the question of what it means for an LLM to code "correctly" for tasks, such as inductive qualitative coding or the interpretation of contested social constructs, where human annotators themselves disagree substantially and where there is no obvious pre-theoretical answer (Dunivin, 2025; Schroeder et al., 2025). Developing standards for validation that are epistemologically appropriate to different research traditions, rather than importing a single positivist metric framework across all use cases, is a significant open agenda for the methodology community.

A second important direction concerns the emergence of agentic LLM workflows, in which models are not simply queried once per document but are embedded in multi-step pipelines that involve tool use, iterative self-critique, web search, code execution, and autonomous decision-making across extended sequences of tasks (Farjam et al., 2025; Schulhoff et al., 2025; Ziems et al., 2024). The workflow described in this paper is deliberately conservative: a human researcher designs a prompt, the LLM processes a batch of documents, and outputs are returned for human validation. This architecture prioritizes transparency and auditability but leaves unexplored the potential of LLMs to participate more actively in the research process. For instance, LLMs could iteratively refine their own coding schemes in response to ambiguous cases, autonomously generate and test alternative interpretations, or orchestrate multi-step information extraction across large and heterogeneous document collections (A. Singh et al., 2024; Farjam et al., 2025).

The methodological implications of such agentic workflows are substantial and largely uncharted. Questions of accountability become more complex when the LLM makes chains of intermediate decisions that are not individually logged or reviewed (Carlson & Burbano, 2025; Schroeder et al., 2025). Validation strategies designed for single-pass annotation may not transfer to multi-step pipelines where errors can compound across stages. And the boundary between the researcher's analytic judgment and the LLM's autonomous contributions becomes harder to locate, raising new questions about transparency and the defensibility of inferences (Roberts et al., 2024; Schroeder et al., 2025; Than et al., 2025). Future methodological work should examine how the principles of human-centered design developed in this paper can be extended to agentic contexts, and what new governance mechanisms are needed to ensure that increased LLM autonomy in the research process does not come at the cost of reproducibility and scholarly accountability (Abdurahman et al., 2025; Barrie, Palmer, et al., 2025; Thomas et al., 2026).

**Limitations**

This paper is intentionally agnostic with respect to specific qualitative and quantitative content analysis traditions, but that breadth comes at the cost of depth. Content analysis spans diverse epistemological commitments and analytic goals (e.g., manifest versus latent content), and no single workflow can substitute for tradition-specific guidance on concept formation, inference, and theorizing (Duriau et al., 2007; Krippendorff, 2012; Reger & Kincaid, 2021). The workflow should therefore be read as cross-paradigmatic guidelines for using LLMs rigorously, to be combined with paradigm-appropriate standards for sampling, interpretive work, and the scope of theoretical claims (Nguyen & Welch, 2025; Schroeder et al., 2025).

The workflow primarily targets prompt-based use of pre-trained LLMs via APIs, because this is currently the most accessible approach for many management researchers. We therefore do not provide detailed guidance on model fine-tuning, domain adaptation, or building custom open-source models, even though these approaches can improve performance and reduce some risks when implemented well (Alizadeh et al., 2024). Likewise, we only briefly touch on advanced architectures and practices such as retrieval-augmented generation, tool-using or "agentic" pipelines, and generative information extraction methods that may be increasingly relevant for organizational research, but which introduce additional design choices and reporting requirements (Xu et al., 2024; Ziems et al., 2024).

A fundamental challenge in LLM-augmented research that the workflow cannot fully resolve is the temporal instability of proprietary models. Commercially deployed LLMs such as GPT and Gemini are updated regularly, and these updates can cause outputs to shift in ways that are difficult to detect and impossible to predict (Abdurahman et al., 2025; Thomas et al., 2026). A research pipeline that yields satisfactory validation metrics at the time of data collection may not produce the same outputs if rerun six months later, even with identical prompts and parameter settings. This creates a replication problem that is qualitatively different from those encountered with traditional computational methods, where algorithms are typically deterministic and inspectable (Thomas et al., 2026). The workflow addresses this risk by recommending that researchers document exact model versions and query dates, retain raw model outputs alongside structured results, and report all parameter settings (Abdurahman et al., 2025; Barrie, Palmer, et al., 2025).

**Conclusion**

This paper provides a comprehensive, practical guide for researchers seeking to integrate LLMs into content analysis workflows. By synthesizing extensive methodological literature and providing detailed supplementary materials including prompt libraries and Python code, it offers a "one-stop shop" for implementing rigorous, transparent, human-centered LLM-augmented research. The workflow emphasizes that while LLMs represent powerful tools for scaling qualitative and quantitative content analysis, they require careful design, continuous human oversight, and thorough validation to produce warrantable knowledge. Researchers who follow these guidelines can harness LLM capabilities while maintaining the methodological rigor essential to quality management research.

# References

A. Singh, A. Ehtesham, S. Kumar, & T. T. Khoei. (2024). Enhancing AI Systems with Agentic Workflows Patterns in Large Language Model. *2024 IEEE World AI IoT Congress (AIIoT)*, 527–532. https://doi.org/10.1109/AIIoT61789.2024.10578990

Abdurahman, S., Salkhordeh Ziabari, A., Moore, A. K., Bartels, D. M., & Dehghani, M. (2025). A Primer for Evaluating Large Language Models in Social-Science Research. *Advances in Methods and Practices in Psychological Science*, *8*(2), 25152459251325174. https://doi.org/10.1177/25152459251325174

Alaei, A. R., Becken, S., & Stantic, B. (2019). Sentiment Analysis in Tourism: Capitalizing on Big Data. *Journal of Travel Research*, *58*(2), 175–191. https://doi.org/10.1177/0047287517747753

Alizadeh, M., Kubli, M., Samei, Z., Dehghani, S., Zahedivafa, M., Bermeo, J. D., Korobeynikova, M., & Gilardi, F. (2024). Open-source LLMs for text annotation: A practical guide for model setting and fine-tuning. *Journal of Computational Social Science*, *8*(1), 17. https://doi.org/10.1007/s42001-024-00345-9

Arora, N., Chakraborty, I., & Nishimura, Y. (2025). AI–Human Hybrids for Marketing Research: Leveraging Large Language Models (LLMs) as Collaborators. *Journal of Marketing*, *89*(2), 43–70. https://doi.org/10.1177/00222429241276529

Ashwin, J., Chhabra, A., & Rao, V. (2025). Using Large Language Models for Qualitative Analysis can Introduce Serious Bias. *Sociological Methods & Research*, 00491241251338246. https://doi.org/10.1177/00491241251338246

Barrie, C., Palaiologou, E., & Törnberg, P. (2025). *Prompt Stability Scoring for Text Annotation with Large Language Models*. https://arxiv.org/abs/2407.02039

Barrie, C., Palmer, A., & Spirling, A. (2025). *Replication for Language Models Problems, Principles, and Best Practices for Political Science*. https://arthurspirling.org/documents/BarriePalmerSpirling_TrustMeBro.pdf

Barros, C. F., Azevedo, B. B., Neto, V. V. G., Kassab, M., Kalinowski, M., Nascimento, H. A. D. do, & Bandeira, M. C. G. S. P. (2025). *Large Language Model for Qualitative Research – A Systematic Mapping Study*. https://arxiv.org/abs/2411.14473

Bermejo, V. J., Gago, A., Gálvez, R. H., & Harari, N. (2025). LLMs outperform outsourced human coders on complex textual analysis. *Scientific Reports*, *15*(1), 40122. https://doi.org/10.1038/s41598-025-23798-y

Bisbee, J., & Spirling, A. (2025). *What to Do When Humans Are No Longer the Gold Standard*. https://github.com/ArthurSpirling/futureProofR

Boonstra, L. (2025). *Prompt Engineering*. https://www.kaggle.com/whitepaper-prompt-engineering

Brandt, P. T., Alsarra, S., D'Orazio, V., Heintze, D., Khan, L., Meher, S., Osorio, J., & Sianan, M. (2025). Extractive versus Generative Language Models for Political Conflict Text Classification. *Political Analysis*, 1–29. Cambridge Core. https://doi.org/10.1017/pan.2025.10027

Bray, T. (2017). *The JavaScript Object Notation (JSON) Data Interchange Format*. https://datatracker.ietf.org/doc/html/rfc8259

Carlson, N., & Burbano, V. (2025). *The Use of LLMs to Annotate Data in Management Research: Foundational Guidelines and Warnings* (SSRN Scholarly Paper No. 4836620). Social Science Research Network. https://doi.org/10.2139/ssrn.4836620

Chae, Y., & Davidson, T. (2025). Large Language Models for Text Classification: From Zero-Shot Learning to Instruction-Tuning. *Sociological Methods & Research*, 00491241251325243. https://doi.org/10.1177/00491241251325243

Chang, K., Xu, S., Wang, C., Luo, Y., Liu, X., Xiao, T., & Zhu, J. (2024). *Efficient Prompting Methods for Large Language Models: A Survey*. https://arxiv.org/abs/2404.01077

Chew, R., Bollenbacher, J., Wenger, M., Speer, J., & Kim, A. (2023). *LLM-Assisted Content Analysis: Using Large Language Models to Support Deductive Coding*. https://arxiv.org/abs/2306.14924

Chollet, F., & Watson, M. (2025). *Deep Learning with Python* (Third Edition). Manning. https://www.manning.com/books/deep-learning-with-python-third-edition

Churchill, A., Pichika, S., Xu, C., & Liu, Y. (2025). GPT models for text annotation: An empirical exploration in public policy research. *Policy Studies Journal*, *n/a*(n/a). https://doi.org/10.1111/psj.70034

Dagdelen, J., Dunn, A., Lee, S., Walker, N., Rosen, A. S., Ceder, G., Persson, K. A., & Jain, A. (2024). Structured information extraction from scientific text with large language models. *Nature Communications*, *15*(1), 1418. https://doi.org/10.1038/s41467-024-45563-x

Dunivin, Z. O. (2025). Scaling hermeneutics: A guide to qualitative coding with LLMs for reflexive content analysis. *EPJ Data Science*, *14*(1), 28. https://doi.org/10.1140/epjds/s13688-025-00548-8

Duriau, V. J., Reger, R. K., & Pfarrer, M. D. (2007). A Content Analysis of the Content Analysis Literature in Organization Studies: Research Themes, Data Sources, and Methodological Refinements. *Organizational Research Methods*, *10*(1), 5–34. https://doi.org/10.1177/1094428106289252

Farjam, M., Meyer, H., & Lohkamp, M. (2025). A Practical Guide and Case Study on How to Instruct LLMs for Automated Coding During Content Analysis. *Social Science Computer Review*, 08944393251349541. https://doi.org/10.1177/08944393251349541

Foisy, L.-O. M., Proulx, É., Cadieux, H., Gilbert, J., Rivest, J., Bouillon, A., & Dufresne, Y. (2025). Prompting the Machine: Introducing an LLM Data Extraction Method for Social Scientists. *Social Science Computer Review*, 08944393251344865. https://doi.org/10.1177/08944393251344865

Friedman, C., Owen, A., & VanPuymbrouck, L. (2025). Should ChatGPT help with my research? A caution against artificial intelligence in qualitative analysis. *Qualitative Research*, *25*(5), 1062–1088. https://doi.org/10.1177/14687941241297375

Garcia Quevedo, D., Glaser, A., & Verzat, C. (2025). Enhancing Theorization Using Artificial Intelligence: Leveraging Large Language Models for Qualitative Analysis of Online Data. *Organizational Research Methods*, 10944281251339144. https://doi.org/10.1177/10944281251339144

Gilardi, F., Alizadeh, M., & Kubli, M. (2023). ChatGPT outperforms crowd workers for text-annotation tasks. *Proceedings of the National Academy of Sciences*, *120*(30), e2305016120. https://doi.org/10.1073/pnas.2305016120

Grimmer, J., Roberts, M. E., & Stewart, B. M. (2022). *Text as Data: A New Framework for Machine Learning and the Social Sciences*. Princeton University Press. https://books.google.co.uk/books?id=dL40EAAAQBAJ

Halterman, A., & Keith, K. A. (2025). Codebook LLMs: Evaluating LLMs as Measurement Tools for Political Science Concepts. *Political Analysis*, 1–17. https://doi.org/10.1017/pan.2025.10017

Hannigan, T. R., Haans, R. F. J., Vakili, K., Tchalian, H., Glaser, V. L., Wang, M. S., Kaplan, S., & Jennings, P. D. (2019). Topic Modeling in Management Research: Rendering New Theory from Textual Data. *Academy of Management Annals*, *13*(2), 586–632. https://doi.org/10.5465/annals.2017.0099

Hayes, A. F., & Krippendorff, K. (2007). Answering the Call for a Standard Reliability Measure for Coding Data. *Communication Methods and Measures*, *1*(1), 77–89. https://doi.org/10.1080/19312450709336664

Hayes, A. S. (2025). "Conversing" With Qualitative Data: Enhancing Qualitative Research Through Large Language Models (LLMs). *International Journal of Qualitative Methods*, *24*, 16094069251322346. https://doi.org/10.1177/16094069251322346

Insch, G. S., Moore, J. E., & Murphy, L. D. (1997). Content analysis in leadership research: Examples, procedures, and suggestions for future use. *The Leadership Quarterly*, *8*(1), 1–25. https://doi.org/10.1016/S1048-9843(97)90028-X

Keraghel, I., Morbieu, S., & Nadif, M. (2024). *Recent Advances in Named Entity Recognition: A Comprehensive Survey and Comparative Study*. https://arxiv.org/abs/2401.10825

Khalid, M. T., & Witmer, A.-P. (2025). *Prompt Engineering for Large Language Model-assisted Inductive Thematic Analysis*. https://arxiv.org/abs/2503.22978

Krippendorff, K. (2012). *Content Analysis: An Introduction to Its Methodology*. SAGE Publications.

LeCun, Y., Bengio, Y., & Hinton, G. (2015). Deep learning. *Nature*, *521*(7553), 436–444. https://doi.org/10.1038/nature14539

Lin, H., & Zhang, Y. (2025). Navigating the Risks of Using Large Language Models for Text Annotation in Social Science Research. *Social Science Computer Review*, 08944393251366243. https://doi.org/10.1177/08944393251366243

Marvin, G., Hellen, N., Jjingo, D., & Nakatumba-Nabende, J. (2024). Prompt Engineering in Large Language Models. In I. J. Jacob, S. Piramuthu, & P. Falkowski-Gilski (Eds.), *Data Intelligence and Cognitive Informatics* (pp. 387–402). Springer Nature Singapore.

McKenny, A. F., Aguinis, H., Short, J. C., & Anglin, A. H. (2018). What Doesn't Get

    Measured Does Exist: Improving the Accuracy of Computer-Aided Text Analysis.

    *Journal of Management*, *44*(7), 2909–2933.

    https://doi.org/10.1177/0149206316657594

Morris, R. (1994). Computerized content analysis in management research: A demonstration

    of advantages & limitations. *Journal of Management*, *20*(4), 903–931.

    https://doi.org/10.1016/0149-2063(94)90035-3

Nguyen, D. C., & Welch, C. (2025). Generative Artificial Intelligence in Qualitative Data

    Analysis: Analyzing—Or Just Chatting? *Organizational Research Methods*,

    10944281251377154. https://doi.org/10.1177/10944281251377154

Ornstein, J. T., Blasingame, E. N., & Truscott, J. S. (2025). How to train your stochastic

    parrot: Large language models for political texts. *Political Science Research and*

    *Methods*, *13*(2), 264–281. Cambridge Core. https://doi.org/10.1017/psrm.2024.64

Pangakis, N., & Wolken, S. (2024). *Keeping Humans in the Loop: Human-Centered*

    *Automated Annotation with Generative AI*. https://arxiv.org/abs/2409.09467

Pangakis, N., Wolken, S., & Fasching, N. (2023). *Automated Annotation with Generative AI*

    *Requires Validation*. https://arxiv.org/abs/2306.00176

Pavlovic, M., & Poesio, M. (2024). The Effectiveness of LLMs as Annotators: A

    Comparative Overview and Empirical Analysis of Direct Representation. In G.

    Abercrombie, V. Basile, D. Bernadi, S. Dudy, S. Frenda, L. Havens, & S. Tonelli

    (Eds.), *Proceedings of the 3rd Workshop on Perspectivist Approaches to NLP*

    *(NLPerspectives) @ LREC-COLING 2024* (pp. 100–110). ELRA and ICCL.

    https://aclanthology.org/2024.nlperspectives-1.11/

Polat, F., Tiddi, I., & Groth, P. (2025). Testing prompt engineering methods for knowledge extraction from text. *Semantic Web*, *16*(2), SW-243719. https://doi.org/10.3233/SW-243719

Reger, R. K., & Kincaid, P. A. (2021). *Content and Text Analysis Methods for Organizational Research*. https://doi.org/10.1093/acrefore/9780190224851.013.336

Roberts, J., Baker, M., & Andrew, J. (2024). Artificial intelligence and qualitative research: The promise and perils of large language model (LLM) 'assistance.' *Critical Perspectives on Accounting*, *99*, 102722. https://doi.org/10.1016/j.cpa.2024.102722

Sahoo, P., Singh, A. K., Saha, S., Jain, V., Mondal, S., & Chadha, A. (2025). *A Systematic Survey of Prompt Engineering in Large Language Models: Techniques and Applications*. https://arxiv.org/abs/2402.07927

Scheuerman, J., & Acklin, D. (2024). A Framework for Enhancing Behavioral Science Research with Human-Guided Language Models. *Proceedings of the AAAI Symposium Series*, *3*(1), 243–247. https://doi.org/10.1609/aaaiss.v3i1.31206

Schmiedel, T., Müller, O., & vom Brocke, J. (2018). Topic Modeling as a Strategy of Inquiry in Organizational Research: A Tutorial With an Application Example on Organizational Culture. *Organizational Research Methods*, *22*(4), 941–968. https://doi.org/10.1177/1094428118773858

Schroeder, H., Aubin Le Quéré, M., Randazzo, C., Mimno, D., & Schoenebeck, S. (2025). Large Language Models in Qualitative Research: Uses, Tensions, and Intentions. *Proceedings of the 2025 CHI Conference on Human Factors in Computing Systems, CHI '25*. https://doi.org/10.1145/3706598.3713120

Schulhoff, S., Ilie, M., Balepur, N., Kahadze, K., Liu, A., Si, C., Li, Y., Gupta, A., Han, H., Schulhoff, S., Dulepet, P. S., Vidyadhara, S., Ki, D., Agrawal, S., Pham, C., Kroiz, G.,

Li, F., Tao, H., Srivastava, A., … Resnik, P. (2025). *The Prompt Report: A Systematic Survey of Prompt Engineering Techniques*. https://arxiv.org/abs/2406.06608

Schwitter, N. (2025). Using large language models for preprocessing and information extraction from unstructured text: A proof-of-concept application in the social sciences. *Methodological Innovations*, *18*(1), 61–65. https://doi.org/10.1177/20597991251313876

Shapiro, G., & Markoff, J. (1997). A Matter of Definition. In *Text Analysis for the Social Sciences*. Routledge.

Stromer-Galley, J., McKernan, B., Zaman, S., Maganur, C., & Regmi, S. (2025). The Efficacy of Large Language Models and Crowd Annotation for Accurate Content Analysis of Political Social Media Messages. *Social Science Computer Review*, 08944393251334977. https://doi.org/10.1177/08944393251334977

Stuhler, O., Ton, C. D., & Ollion, E. (2025). From Codebooks to Promptbooks: Extracting Information from Text with Generative Large Language Models. *Sociological Methods & Research*, *54*(3), 794–848. https://doi.org/10.1177/00491241251336794

Than, N., Fan, L., Law, T., Nelson, L. K., & McCall, L. (2025). Updating "The Future of Coding": Qualitative Coding with Generative Large Language Models. *Sociological Methods & Research*, *54*(3), 849–888. https://doi.org/10.1177/00491241251339188

Thomas, L. D. W., Romasanta, A. K. G., & Pujol Priego, L. (2026). Jagged competencies: Measuring the reliability of generative AI in academic research. *Journal of Business Research*, *203*, 115804. https://doi.org/10.1016/j.jbusres.2025.115804

Törnberg, P. (2024a). Best Practices for Text Annotation with Large Language Models. *Sociologica*, *18*(2), 67–85. https://doi.org/10.6092/issn.1971-8853/19461

Törnberg, P. (2024b). Large Language Models Outperform Expert Coders and Supervised Classifiers at Annotating Political Social Media Messages. *Social Science Computer Review*, 08944393241286471. https://doi.org/10.1177/08944393241286471

Tseng, Y.-M., Chen, W.-L., Chen, C.-C., & Chen, H.-H. (2025). *Evaluating Large Language Models as Expert Annotators*. https://arxiv.org/abs/2508.07827

Vaswani, A., Shazeer, N., Parmar, N., Uszkoreit, J., Jones, L., Gomez, A. N., Kaiser, L., & Polosukhin, I. (2017). *Attention Is All You Need*. https://arxiv.org/abs/1706.03762

Watkins, R. (2024). Guidance for researchers and peer-reviewers on the ethical use of Large Language Models (LLMs) in scientific research workflows. *AI and Ethics*, *4*(4), 969–974. https://doi.org/10.1007/s43681-023-00294-5

Wei, J., Tay, Y., Bommasani, R., Raffel, C., Zoph, B., Borgeaud, S., Yogatama, D., Bosma, M., Zhou, D., Metzler, D., Chi, E. H., Hashimoto, T., Vinyals, O., Liang, P., Dean, J., & Fedus, W. (2022). Emergent Abilities of Large Language Models. *Transactions on Machine Learning Research*. https://openreview.net/forum?id=yzkSU5zdwD

Wetzels, M., Wetzels, R., Schweiger, E., & Grewal, D. (2025). How Topic Modeling Can Spur Innovation Management. *Journal of Product Innovation Management*, *42*(5), 921–946. https://doi.org/10.1111/jpim.12790

Xu, D., Chen, W., Peng, W., Zhang, C., Xu, T., Zhao, X., Wu, X., Zheng, Y., Wang, Y., & Chen, E. (2024). Large language models for generative information extraction: A survey. *Frontiers of Computer Science*, *18*(6), 186357. https://doi.org/10.1007/s11704-024-40555-y

Yang, E., Wang, Z., Zhou, C., & Xu, Y. (2025). *Data Annotation with Large Language Models: Lessons from a Large Empirical Evaluation*.

Ziems, C., Held, W., Shaikh, O., Chen, J., Zhang, Z., & Yang, D. (2024). Can Large Language Models Transform Computational Social Science? *Computational Linguistics*, *50*(1), 237–291. https://doi.org/10.1162/coli_a_00502

# Appendix A – Sample prompt

Below you can find a simple sample prompt that demonstrates three content analysis tasks (annotation, summarization and information extraction) on analyzing an academic paper. An elaborated version of this prompt could be used for a literature review. Each variable instructs LLM to perform a specific task, identified by the prefix: "AN_" for annotation, "SU_" for summarization and "IE_" for information extraction. The prompt uses simple role specification (aka persona prompting) and zero-shot instructions for each task.

**PROMPT**

You are a researcher in Management. Please use the following coding manual to code the text.

- IE_YEAR_PUBLISHED = The year the paper was published.
- IE_JOURNAL = Journal in which the paper was published.
- IE_AUTHORS = Authors of the paper.
- IE_TITLE = Title of the paper.
- IE_DOI = DOI of the paper.
- AN_RQ = Did the authors explicitly define research questions, research aims, or research objectives? (1 = Yes, 0 = No).
- SU_RQ_VERBATIM = Copy verbatim text from the paper that defines the research questions/aims/objectives. If none, return N/A.
- SU_FULL_PROCESS = What were the steps of the research and analysis process from start to finish?
- SU_SUMMARY = Summarize the paper in one paragraph.
- SU_KEY_TAKEAWAYS = What are the key takeaways of the paper?
- AN_PAPER_TYPE = Type of the paper – empirical, review, conceptual, editorial, other.
- SU_PAPER_TYPE_REASON = Explain why you chose the paper type.
- AN_METHOD_TYPE = What type of research methods were used (qualitative, quantitative, multiple, mixed)?
- SU_METHOD_TYPE_REASON = Explain why you chose the research method type.
- SU_METHOD = Describe research methods that were used in the paper.
- SU_DATA_COLLECTION = How was the data collected?
- SU_DATA_COLLECTION_VERBATIM = Copy verbatim text from the paper that describes data collection. If there are multiple relevant sections combine them (separated by "…") in a single text. Return N/A if none.
- IE_SAMPLE_SIZE = What was the size of the sample? Return N/A if not explicitly stated.
- SU_KEY_FINDINGS = What were the key findings of the paper?
- AN_THEORETICAL_CONTRIBUTION = Was the theoretical contribution explicitly described? (1 = Yes, 0 = No).
- SU_THEORETICAL_CONTRIBUTION_VERBATIM = Copy verbatim text from the paper that describes the theoretical contribution of the study. If there are multiple relevant sections combine them (separated by "…") in a single text. Return N/A if none.

# Output Format
Return **ONLY** a JSON array of objects. Do not include markdown formatting like ```json or intro text. Use this exact schema:

```
[
{
 "IE_YEAR_PUBLISHED": "string",
 "IE_JOURNAL": "string",
 "IE_AUTHORS": "string",
 "IE_TITLE": "string",
 "IE_DOI": "string",
 "AN_RQ": "integer",
 "SU_RQ_VERBATIM": "string",
 "SU_FULL_PROCESS": "string",
 "SU_SUMMARY": "string",
 "SU_KEY_TAKEAWAYS": "string",
 "AN_PAPER_TYPE": "string",
 "SU_PAPER_TYPE_REASON": "string",
 "AN_METHOD_TYPE": "string",
 "SU_METHOD_TYPE_REASON": "string",
 "SU_METHOD": "string",
 "SU_DATA_COLLECTION": "string",
 "SU_DATA_COLLECTION_VERBATIM": "string",
 "IE_SAMPLE_SIZE": "string",
 "SU_KEY_FINDINGS": "string",
 "AN_THEORETICAL_CONTRIBUTION": "integer",
 "SU_THEORETICAL_CONTRIBUTION_VERBATIM": "string"
}
]
```

# Appendix B – Supplementary code description and instructions

We provide Python code in the form of Jupyter notebooks to assist researchers implement the workflow. The notebooks use Open AI or Google Gemini API to process text data stored in a Google Drive folder. API Keys need to be acquired from the provider before calling API to process text. Researchers can run the code in Google Colab online environment. The code effectively loops through pdf files in the specified folder, sends each one to the LLM for processing, and writes the results in the output Excel file. The process can be applied to any text data as long as it is provided in the form of pdf files. The only things researchers need to change in the code are (1) the prompt, (2) structure of the output file and (3) destinations of input/output files.

The Jupyter notebook has four blocks, each can be run separately.

1. Set up the connection to the LLM and Google Drive. This block uses API Key set up in Colab secret variable.
2. Set up folders for input/output files. Create output and processed list Excel files. Set up output table structure. This is the only block that needs changing when the process is applied on new data.
3. Process all files in the designated Google Drive folder and write the results in the output Excel file.
4. Show the results in the output Excel file.

The code is available on this link: https://github.com/IvanZupic/LLMWorkflow